\documentclass[10pt,twocolumn,letterpaper]{article}

\usepackage[pagenumbers]{cvpr} 
\usepackage{threeparttable}
\usepackage{amsmath}
\usepackage{multirow}
\usepackage{etoolbox}
\usepackage{amssymb} 
\definecolor{cvprblue}{rgb}{0.21,0.49,0.74}
\usepackage[pagebackref,breaklinks,colorlinks,allcolors=cvprblue]{hyperref}

\def\paperID{17738} 
\def\confName{CVPR}
\def\confYear{2025}

\title{GaussianProperty: Integrating Physical Properties to 3D Gaussians with LMMs}

\author{%
Xinli Xu$^{1}$\thanks{Equal contribution.}  \quad Wenhang Ge$^{1*}$ \quad Dicong Qiu$^{1*}$ \quad ZhiFei Chen$^{1}$  \quad Dongyu Yan$^{1}$ \quad Zhuoyun LIU$^{1}$ \quad \\
Haoyu Zhao$^{1}$ \quad 
Hanfeng Zhao$^{3}$     \quad Shunsi Zhang$^{3}$  \quad Junwei Liang$^{1,2}$ \quad Ying-Cong Chen$^{1,2}$\thanks{Corresponding author.}  \quad  \vspace{8pt}\\
HKUST(GZ)\textsuperscript{1} \quad HKUST\textsuperscript{2} \quad Quwan\textsuperscript{3}}

\begin{document}
\maketitle
\begin{abstract}
Estimating physical properties for visual data is a crucial task in computer vision, graphics, and robotics, underpinning applications such as augmented reality, physical simulation, and robotic grasping. However, this area remains under-explored due to the inherent ambiguities in physical property estimation.
To address these challenges, we introduce \textbf{GaussianProperty}, a training-free framework that assigns physical properties of materials to 3D Gaussians. Specifically, we integrate the segmentation capability of SAM with the recognition capability of GPT-4V(ision) to formulate a global-local physical property reasoning module for 2D images. Then we project the physical properties from multi-view 2D images to 3D Gaussians using a voting strategy.
We demonstrate that 3D Gaussians with physical property annotations enable applications in physics-based dynamic simulation and robotic grasping.
For physics-based dynamic simulation, we leverage the Material Point Method (MPM) for realistic dynamic simulation.
For robot grasping, we develop a grasping force prediction strategy that estimates a safe force range required for object grasping based on the estimated physical properties.
Extensive experiments on material segmentation, physics-based dynamic simulation, and robotic grasping validate the effectiveness of our proposed method, highlighting its crucial role in understanding physical properties from visual data. Online demo, code, more cases and annotated datasets are available on the project page: \href{https://Gaussian-Property.github.io}{https://Gaussian-Property.github.io}

\end{abstract}

\begin{figure}
\begin{center}
\includegraphics[width=1\linewidth]{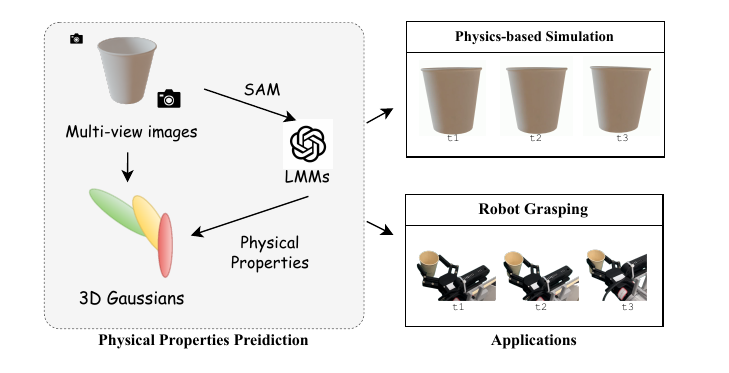}
\end{center}
   \caption{
   \textbf{GaussianProperty} 
   is a training-free framework, aiming at adding physical properties to 3D Gaussians with the assistance of LMMs. By assigning physical properties to 3D Gaussians, it promotes several downstream tasks such as physical-based generative dynamics and robot grasping in this work.} 
\label{fig:teaser}
\end{figure}

\section{Introduction}

Estimating physical properties from visual data is a critical task in both computer vision and graphics, serving as the foundation for various fields, including augmented reality (AR) \cite{imbert2013adding, alam2021factors, azuma1997survey}, robotic grasping \cite{caldera2018review, bicchi2000robotic, saxena2008robotic}, and physics-based dynamic simulation \cite{carlson1996don, goovaerts2000estimation, hu2019difftaichi}. Recently, the integration of physical properties into 3D model has generated significant interest across these domains, underscoring the need for precise physical property estimation. However, this area remains under-explored due to the inherent ambiguities in physical property estimation. Key challenges include the difficulty of acquiring labeled ground-truth data, as intrinsic physical properties are not directly observable through visual means, and the ambiguity of the prediction task, which is further compounded by the limited number of observable surfaces.

Humans possess a remarkable ability to predict the physical properties of objects based on visual cues alone \cite{mcadams1993recognition}. Research in cognitive science and human vision suggests that this capability stems from our skill in associating visual appearances with previously encountered materials, about which we have developed a rich and grounded understanding. This process allows us to intuitively gauge physical property such as weight, texture, and density from visual observation.
Recently, Large Language Models (LLMs) have achieved impressive progress in nature language understanding. Based on this, Large Multimodal Models (LMMs) extend LLMs by further incorporating image modality into the model training. With a massive repository of prior knowledge, which covers the task of physical property estimation, showcasing a robust understanding and recognition capabilities of visual data that mirrors human perception. We show an example in Figure \ref{fig:gpt}. 

In this study, we introduce a novel method called \textit{GaussianProperty}, designed to assign physical properties to 3D model (i.e., 3D Gaussians) using Segment Anything (SAM) and GPT-4V. We demonstrate that incorporating physical properties into 3D model enhances two downstream tasks: physics-based dynamic simulation and robotic grasping.
For physics-based dynamic simulation, we leverage a custom Material Point Method (MPM) to enrich 3D Gaussians with physical properties estimated from multi-view 2D images, enabling realistic dynamic simulation.
For robotic grasping, we develop a grasping force prediction module. Based on the estimated physical properties of 3D model, this module predicts the upper bound force to avoid object deformation and the lower bound force required to lift the object without slipping, ensuring proper grasping force estimation. 

Specifically, we leverage the recognition capabilities of GPT-4V to estimate physical properties from 2D images. However, predicting properties for complex scenes containing multiple components with distinct physical characteristics from a single global image presents significant challenges. To address this, we first use the robust segmentation capabilities of SAM \cite{kirillov2023segment} to segment each component within the global image. We then employ GPT-4V, incorporating both global and detailed local information from each segmented part and its spatial context, to achieve more precise physical property estimations. 

After acquiring physical properties from 2D images, we project this information onto 3D Gaussians using a multi-view reconstruction approach and a voting strategy.
The 3D Gaussians, representing an explicit 3D point cloud format, support effective reconstruction from multi-view images.
To be specific, we first reconstruct the 3D Gaussian representation using multi-view images. We then project the spatial positions of the 3D Gaussians onto the visible 2D images to gather corresponding estimations. A voting strategy is subsequently employed to determine the final physical properties of the 3D Gaussians, effectively avoiding occasional errors that may occur in a single view.

For robotic grasping, we select some common objects from daily life to validate the effectiveness of the adaptively adjusted grasping force predicted by physical properties. We compare the grasping success ratio and deformation ratio with those obtained using a fixed force.


To summarize, our contributions are listed as follows.
\begin{itemize}
    \item We present the first exploration of leveraging Large Multimodal Models (LMMs), e.g. GPT-4V for physical property estimation for 3D model, showing robust results in physical properties estimation.
    \item We demonstrate two crucial downstream tasks that benefits from estimated physical properties, i.e., physical-based dynamic simulation and robotic grasping. 
    \item Extensive experiments including materials segmentation, realistic dynamic simulation and real-world grasping validate the effectiveness of our proposed method, showing superior performance and benefiting downstream tasks.
\end{itemize}

\section{Related Work}

\subsection{Physical property estimation for 3D models}
In the burgeoning field of 3D modeling, the accurate estimation of physical properties such as density, elasticity, and thermal conductivity is a long-standing problem \cite{adelson2001seeing, wu2015galileo}, serving critical roles in downstream tasks like AR, robotics, and physical-based simulation. Although promising, existing work mostly tackles specific types of material properties, e.g. mass or tenderness, by collecting corresponding task-dependent data with little generalization. In contrast, our method can generate diverse physical properties like mass density, friction, and hardness in a zero-shot manner with the recognition capability of LLMs. Several works have explored LLMs for physical property estimation. For example, NeRF2Physics \cite{zhai2024physical} leverages large language models to propose candidate materials for objects, constructing a language-embedded point cloud to estimate physical properties such as mass, friction, and hardness through a zero-shot kernel regression approach. Make-it-real \cite{fang2024make} reasons the PBR materials including albedo, metallic, and roughness for 3D assets texture generation.

\subsection{Multimodal Large Language Models.}
Large Language Models (LLMs) have achieved impressive progress in recent years, demonstrating a strong capability in understanding natural language. However, they generally lack the ability to reason about images, as they lack image priors for training. With the growing demand for this capability, recent research has focused on developing Large Multimodal Models (LMMs) that integrate image modalities for training. The state-of-the-art models \cite{brown2020language,openai2023gpt4v,team2023gemini,touvron2023llama,liu2023llava} have been leveraged in various downstream applications, such as image captioning \cite{luo2024scalable}, physically based rendering (PBR) materials estimation \cite{fang2024make}, and 3D grounding \cite{wang2023chat}. LMMs have shown great potential for these tasks, significantly improving performance. The introduction of GPT-4V \cite{openai2023gpt4v} has notably advanced the capabilities of large multimodal models, showcasing exceptional 2D comprehension and extensive open-world knowledge. While GPT-4V is not designed to process 3D data directly, the innovative GPTEval3D \cite{wu2024gpt4vision} has successfully utilized GPT-4V to assess the quality of 3D objects, finding that its evaluations closely match those of humans.  Additionally, other models such as BLIP-2 \cite{li2023blip} and Flamingo \cite{alayrac2022flamingo} have further pushed the boundaries of image-text understanding and generation, offering new possibilities for multimodal research and applications. The continual evolution of LMMs promises to drive further advancements in fields requiring integrated image and text reasoning capabilities.

\subsection{Dynamic Rendering}
Neural Radiance Fields (NeRF) \cite{mildenhall2021nerf} have garnered significant interest in recent years due to their remarkable capabilities in multi-view 3D reconstruction. An evolutionary advancement within the NeRF framework is the incorporation of a temporal dimension, enhancing the representation of dynamic scenes. For instance, D-NeRF \cite{pumarola2021d} and NeRFies \cite{park2021nerfies} have extended time-dependent neural fields by decomposing them into an inverse displacement field and canonical time-invariant neural fields.
Furthermore, 3D Gaussian splatting \cite{kerbl20233d}, a point-based rendering technique, has gained popularity for its highly realistic rendering quality and efficient training speed. Building on this, Dynamic 3D Gaussians \cite{luiten2023dynamic, wu20234d} have successfully integrated the temporal dimension to more effectively represent dynamic scenes. However, existing methods for dynamic rendering typically rely on video sequences for supervision, where the 3D models are deformed to align consistently with the video footage.
In this study, since we assign the physical  properties for 3D Gaussians, we assist dynamic simulations seamlessly integrate the simulation within the GS framework.

\subsection{Material-sensitive Robot Grasping}

Soft robotic grippers \cite{li2019vacuum, zaidi2021actuation, shintake2018soft} leverage the deformation and compliance properties of soft materials enabling grippers to automatically adapt to the geometries and various weights of the objects being grasped. This adaptability necessitates the careful selection of materials and mechanical designs tailored to specific applications, limiting the generality of such solutions across all scenarios.
Optical tactile sensing approaches \cite{liu2022gelsight, jiang2021finger, lepora2021soft} requires a camera positioned within each fingertip of a gripper, situated behind a soft and transparent artificial skin, to convert optical observations of markers printed on the skin to force estimations; while electronic skins \cite{qiu2022nondestructive, shih2020electronic} detects exerted forces from electric signals. However, these two approaches often face challenges related to durability, and some require significant additional installation space, limiting their practicality in certain applications.
In this work, we propose integrating \textit{GaussianProperty} to enable material-sensitive robot grasping, which takes merely the visual inputs from a camera to predict the composing materials and estimate corresponding physical properties of the object to grasp. Our approach can be easily adapted to a wide spectrum of robotic and industrial applications.

\begin{figure*}
\centering
\includegraphics[width=0.9\textwidth]{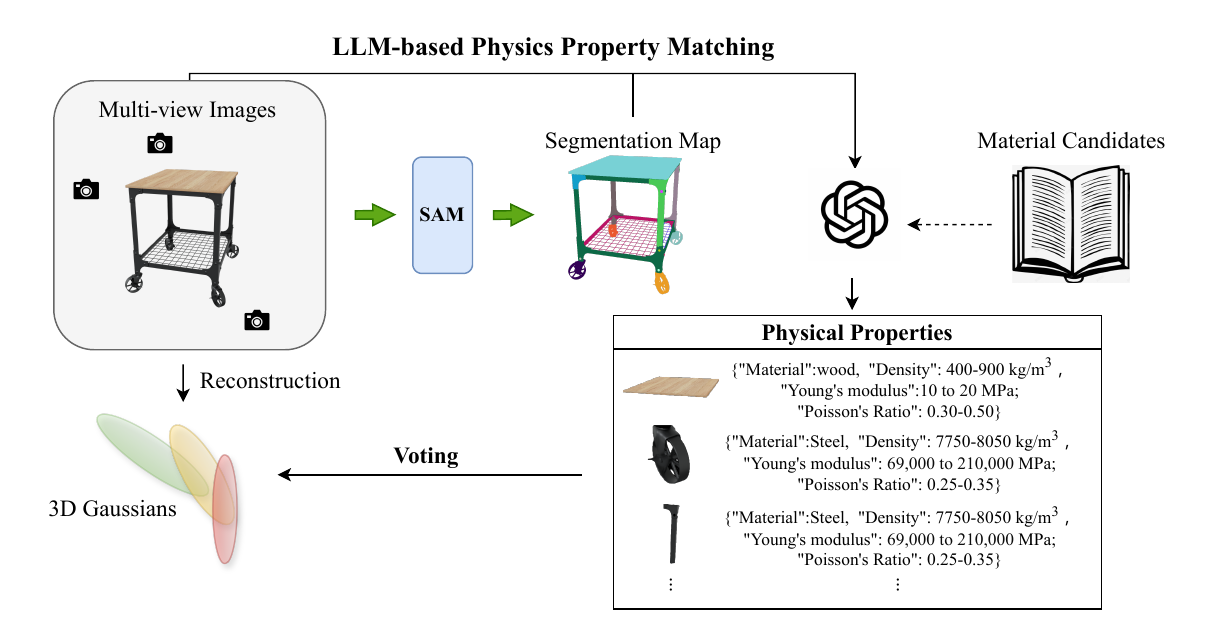}

   \caption{\textbf{Overall pipeline}. Our Gausssian-Property initially leverages SAM to get the segmentation map of the object. Then the original images and the masks are sent to the foundation 
    models like GPT-4V(ision) to get the corresponding physical properties by inquiring the material candidates. After acquiring physical properties from 2D images, we using a multi-view approach and a voting strategy to add physical properties to the reconstruction 3D Gaussians.} 
\label{fig:framework}
\end{figure*}


\section{Method}

\subsection{Problem Formulation}

Given a well-reconstructed 3D Gaussian representation, our objective is to attribute physical properties to each Gaussian. The specific physical property can vary according to the downstream task. 
In this work, we demonstrate a potential application in physics-based dynamic simulation via Material Point Method (MPM) and robotic grasping. The former application  requires material density $\rho$, Young's modulus $E$, Poisson's ratio $P$, and material type $T$. And robotic grasping requires the material density $\rho$, volume $V$, friction coefficient $\mu$, thickness $d$, maximal tolerable curvature $\kappa$, Young's modulus $E$. 
An overview of our framework is illustrated in Figure \ref{fig:framework}.

\subsection{Part-Level  Segmentation}

Understanding an object's physical properties requires delving into the characteristics of its individual parts, as each part may present unique attributes. Considering this, we utilize SAM for image segmentation, adeptly predicts masks with precise boundaries that capture whole, part, and subpart levels, thereby reflecting the object's hierarchical semantic structure.
In this work, we emphasize the significance of part-level information, which enables us to dissect an object into its constituent parts. This approach facilitates a more accurate and exhaustive comprehension of the physical properties of visual data.
Our method not only harnesses the semantic stratification provided by SAM but also actively integrates it to remedy the ambiguity arising from objects possessing multiple physical attributes.

Concretely, for each image \(I\) within the observed set $\mathcal{I}^{N}$ , we input a grid of $32 \times 32$ point prompts. SAM responds by segmenting precise masks at varying levels based on the prompts at these points. We operate using the part-level semantic mask $\mathbf{M}$, subsequently refining the segmentation by eliminating superfluous masks within each of the three mask sets. This culling is informed by predicted intersection-over-union (IoU) scores, stability scores, and the overlap rates between masks. The resulting segmentation maps meticulously trace the boundaries of objects at their respective hierarchical levels, effectively segmenting the scene into semantically coherent regions.

\subsection{Physics Property Matching}
After achieving precise part-level semantic segmentation, the next step is to match the segmented parts with their corresponding physical properties, a process we term Physics Property Matching. We discussed the establishment of material candidates in Section~\ref{material_candidate} and utilizing a combination of global and local knowledge in Section~\ref{global-local} to assist GPT-4V in recognizing the material properties of the object. Additionally, we discuss the Gradual Prompt Guidance in Section~\ref{guidance} to help the model progressively build an understanding of the entire object and discern the association between its parts and the whole.

\subsubsection{Material Candidates} \label{material_candidate}
Our approach leverages a curated collection of candidate materials, consisting of fifteen ubiquitous material families and more than 600 materials, integral to everyday objects and structures. This library encompasses a wide range of materials, ensuring comprehensive coverage of various densities and material properties.
The common object material library includes density ranges for a variety of materials. For instance, metals such as aluminum (2700 kg/m³), steel (7750-8050 kg/m³), and copper (8920-8960 kg/m³) are covered, as well as non-metals like glass (2200-2500 kg/m³), concrete (2300-2500 kg/m³), and plastics such as polyethylene (930-970 kg/m³). This diversity highlights the extensive range of physical properties found in commonly used substances.

This robust material database is the cornerstone of our physical property matching process. By offering a comprehensive material library, the material candidates simplify material retrieval for the LLM model. Additionally, it minimizes ambiguity in property predictions from different perspectives, ensuring accuracy. Reliable material identification thus provides a dependable reference.

\subsubsection{ Combined Global-Local Reasoning Module } \label{global-local}

\begin{figure*}
\centering
\includegraphics[width=0.9\textwidth]{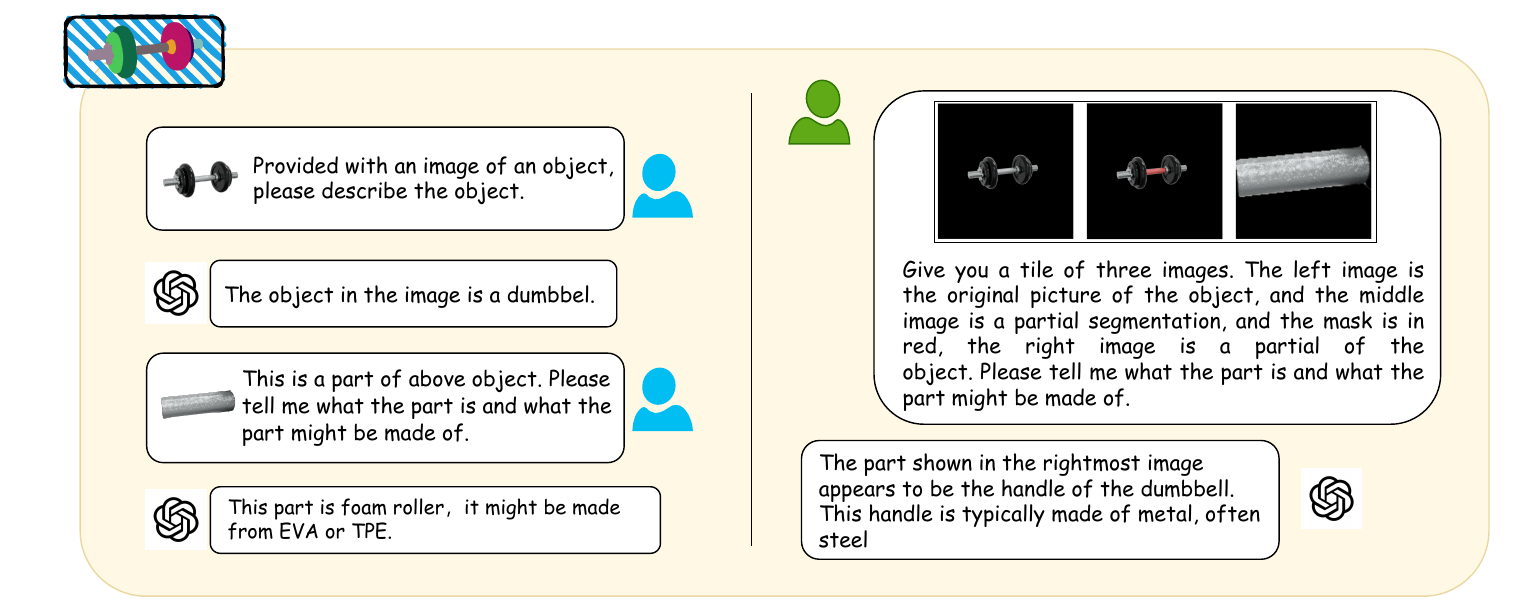}

   \caption{\textbf{Left}: GPT-4V(ision) struggles to recognize the material when directly provided with both global and partial image inputs. \textbf{Right}: Enhanced with combined global-local information and association, the agent accurately characterizes the component's properties.} 
\label{fig:gpt}
\end{figure*}

Our observation revealed that utilizing a global-to-local knowledge framework significantly improves the accuracy in assigning physical properties to each part. A straightforward method involves having the model understand the entire object first and then evaluate a part of the object. However, we found it challenging for the model to establish a connection between the whole and its parts, as shown in Figure \ref{fig:gpt} \textbf{(Left)}. Motivated by this insight, we built a bridge between global and local information, enabling the model to understand their connection. As shown in Figure \ref{fig:gpt} \textbf{(Right)}, the left image displays the original object, the middle image shows a partial segmentation with the mask highlighted in red, and the right image depicts a specific part of the object. Starting from this global perspective, GPT-4V then focuses on the details of each part, incorporating local cues such as texture, color, and contextual information from adjacent parts. This approach aids in accurately identifying each part and inferring its material composition.

\subsubsection{Gradual Prompt Guidance} \label{guidance}

We design gradual prompt guidance to help the LMMs gradually build an understanding of the entire object and then discern the association between its parts and the whole through the segment map. The prompt instructs the LLM to first briefly describe the part based on the provided image and then identify the material of the part, specifying its mass density, Young's modulus, and Poisson's Ratio. The material types are selected from a predefined  material candidates of common object. This structured approach ensures that the LMMs can effectively comprehend the context and specifics of each part, thereby enhancing its accuracy in identifying physical properties. The "Gradual Prompt Guidance" design thus provides a systematic method to improve the model's understanding and performance by leveraging both global and local information. 

\subsection{Lift 2D to 3D via Voting}
\subsubsection{3D  Reconstruction from Multi-view Images}

3D Gaussian Splatting method has the advantage of providing an explicit 3D representation, making it easy to add any other properties. 
This method reparameterizes NeRF with a set of unstructured 3D Gaussian kernels \(\{x_p, \sigma_p, A_p, C_p\}_{p \in P}\), where \(x_p\), \(\sigma_p\), \(A_p\), and \(C_p\) denote the centers, opacities, covariance matrices, and spherical harmonic coefficients of the Gaussians, respectively.
A differentiable rasterization rendering method is employed to project 3D Gaussians to 2D images to compare the rendered image with ground-truth image by
\[ C = \sum_{k \in P} \alpha_k \text{SH}(d_k; C_k) \prod_{j=1}^{k-1} (1 - \alpha_j), \]
where \(\alpha_k\) are the z-depth ordered opacity, and \(d_k\) is the view direction from the camera to \(x_k\). 

\subsubsection{Frequency-based Voting Strategy}

Through reconstruction, we obtain 3D Gaussians denoted as $GS$. Previous works \cite{qin2023langsplat, zhou2023feature} incorporate CLIP features into 3D Gaussians through training, but the process is time-consuming and scene-specific, limiting downstream applications.
Alternatively, we lift the 2D information to 3D models with a projection based method. Each 3D Gaussian \(\mathbf{s} \in GS\)  is projects to each 2D image \(I \in \mathcal{I}^{N}\), we determine the pixel coordinates \((u, v)\) on 2D plane using the camera parameters. The projection is performed as 
\begin{equation}
    u, v =
\mathbf{K} \cdot [\mathbf{R} | \mathbf{t}] \cdot 
\mathbf{s},
\end{equation}
where \(\mathbf{K}\) is the camera intrinsic matrix, \([\mathbf{R} | \mathbf{t}]\) represents the rotation and translation matrices (extrinsic parameters), and $ \mathbf{s}$ is the coordinates of the  point.

However, the projected pixel coordinates are meaningless if the point is invisible in the source image. Thus, We estimate the visibility using the Gaussian-estimated depth to determine if the point is visible of the image. The voting strategy involves projecting each Gaussian to all the visible views and retrieving the corresponding properties. 
To ensure consistency across multi-view images, we adopt a frequency-based voting strategy. The attribute with the highest frequency is chosen as the final predicted attribute. The voting process can be described as:

\[
\hat{a} = \arg\max_{a} \sum_{i=1}^{N} \mathbb{I}(a_i = a),
\]
where \(\hat{a}\) is the predicted attribute, \(N\) is the number of views, \(a_i\) is the attribute observed in the \(i\)-th view, and \(\mathbb{I}\) is the indicator function that equals 1 if the attribute matches and 0 otherwise.


\subsection{ Material-sensitive Robot Grasping}

The diversity of objects in the real world, composed of various materials and physical properties, makes it impractical to use a single grasping force for all. An adaptive strategy is essential to calibrate the grasping force according to the specific materials of the object being manipulated. The grasping force applied by the robotic gripper must be sufficient to lift the target object without slipping while remaining below a threshold to prevent damage or deformation.
These two criteria effectively define the lower bound $F_{\min}$ and the upper bound $F_{\max}$ of the grasping force $F$.
\begin{align*}
& F_{\min} \leq F \leq F_{\max} \\
& F_{\min} = \sum_{i=1}^{M} \frac{1}{2} \rho(i) V(i) g \left( \frac{\cos\theta}{\mu(s) }- \sin\theta \right) \\
& F_{\max} = \min\left[ A \sigma_{y}(s), \frac{1}{2} A E(s) d(s) \kappa_{\max}(s) \right]
\end{align*}
where the object consists of $M$ parts, and even physical property distribution of material is assumed within each part; $s \in \{ 1, \cdots, M \}$ refers to the object part containing the force bearing surface; $\rho(\cdot)$ and $V(\cdot)$ are respectively the density and the volume of a part; $\theta$ is the lifting angle of the gripper; $\mu(\cdot)$ is the friction coefficient between the gripper tips and a surface; $A$ is the area of a force bearing surface; $d(\cdot)$ is the thickness of a surface, $\kappa_{\max}(\cdot)$ is maximal tolerable curvature of a surface; $E(\cdot)$ is Young's modulus of electricity of the material of a part; and $g \approx 9.8 m/s^{2}$ is the gravity constant. Specific values of $\rho$, $\mu$ and $E$ relate directly to the predicted material of each part, while those of $V$ and $d$ can be estimated from object reconstruction, and $A$ is approximated with the area of the gripper finger tips.

To maximize the grasping reliability, confining the grasping force within the robotic gripper capability, and attempting to avoid the gripper executing commands close to its input bounds with $0 \leq \eta \leq 1$ margin, an optimal choice of grasping force 
\[
F^{*} = \begin{cases}
    \left[ \bar{F} \right]_{\left[ F_{\min} \right]_{G} + \eta \Delta F}^{\left[ F_{\max} \right]_{G} - \eta \Delta F} & F_{\min} < F_{\max} \\
    \left[ \bar{F} \right]_{G} & F_{\min} \geq F_{\max}
\end{cases} 
\]
with $\left[ \cdot \right]_{G}$ and $\left[ \cdot \right]_{\min}^{\max}$ clipping a force within the input range of robotic gripper $G$ and between some lower and upper bounds. $ \Delta F = \max\left[ 0, \left[ F_{\max} \right]_{G} - \left[ F_{\min} \right]_{G} \right]$. And $F^{*}$ remains optimality in extreme situations where $F_{\min} > F_{\max}$. See the Supplementary for detailed derivation.

\subsection{Physics-based Dynamic Simulation}

Previous works, such as PhysGaussian~\cite{xie2023physgaussian}, have achieved dynamic simulation by integrating Newtonian physics directly into 3D Gaussian representations, using the Material Point Method (MPM) to enable realistic physical interactions. MPM combines the strengths of both particle simulation methods and grid-based finite element methods (FEM) to effectively handle complex problems involving large deformations, phase changes, and interactions between multiple materials. However, a key limitation in these approaches is the need for manual assignment of physical properties to each Gaussian point, such as  material type and physical properties corresponding to the material. This manual assignment is time-consuming and not realistic.

To address this inefficiency, our method can directly predicts the physical properties of each Gaussian point, thus eliminating the need for manual assignment. Specifically, we employ a combination of multi-view 2D-to-3D projection and frequency-based voting to derive these properties from observed images. For each Gaussian point in the 3D representation, our model predicts essential physical attributes, including density (\(\rho\)), Young's modulus (\(E\)), Poisson’s ratio (\(P\), among others. This prediction process begins with segmenting observed images at the part level to ensure each segment’s unique physical characteristics are represented accurately. We then apply a voting strategy to integrate physical properties across multiple views, ensuring consistency and robustness in the 3D representation. By automating the assignment of these properties through \textit{GaussianProperty}, we significantly reduce the time required for dynamic simulations, streamline the simulation workflow, and enable scalable applications in complex environments. We show some cases in Figure \ref{fig:dynamic}.

\begin{table*}[ht]
    \small
    \renewcommand\arraystretch{1}
    \setlength\tabcolsep{1.0pt}
\caption{Comparison of material segmentation with NeRF2Physics \cite{zhai2024physical} across different categories on ABO and MVImgNet  dataset. Our method achieves a more comprehensive and accurate understanding of the object and achieve more precise material segmentation.}
\vspace{-0.4cm}
\begin{center}
\begin{tabular}{l|cccccc|cccccccccccc}
			\toprule[1pt]
·			\multirow{2}*{Method}  & \multicolumn{6}{c|}{ABO dataset} & \multicolumn{11}{c}{MVImgNet }\\ 
			\cline{2-18}
                & Wood & Metal & Plastic & Fabric & Ceramic & Average & Wood & Metal& Plastic & Glass & Fabric & Foam & Food & Ceramic & Paper & Leather & Average\\
                \midrule
			Nerf2phycics & 27.87 & 13.01 & 8.38 & 40.26 &38.44& 25.59 & 6.39 & 3.63 & 6.70 & 1.15 & 1.11 & 0.38 & 2.40 & 6.54 & 6.73 & 5.20 & 4.02 \\
			Ours & \textbf{61.53} &\textbf{33.41} & \textbf{38.26} & \textbf{67.57} & \textbf{78.40} & \textbf{55.83} & \textbf{41.96} & \textbf{38.85} &\textbf{ 39.50} & \textbf{18.87} & \textbf{27.12}&\textbf{23.18} &\textbf{ 84.89} &\textbf{ 19.74} &\textbf{ 30.23 }& \textbf{23.96} & \textbf{34.83}\\
            \bottomrule[1pt]
		\end{tabular}
	\end{center}
	\label{table:material}
\end{table*}

\begin{figure*}
\begin{center}
\includegraphics[width=0.9\textwidth]{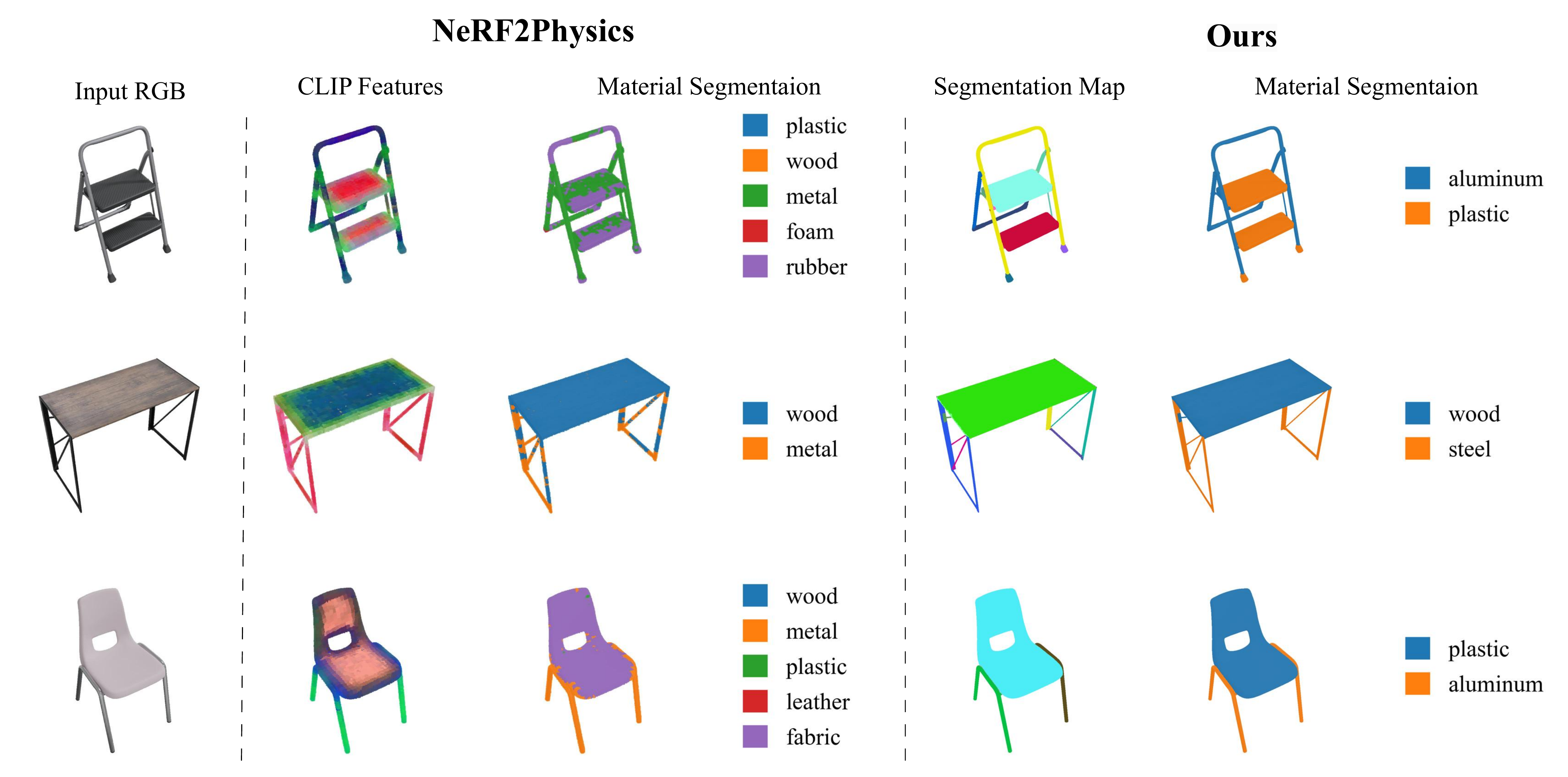}
\end{center}
\vspace{-0.4cm}
   \caption{\textbf{Qualitative results of Material Segmentation}. Our model makes  boundary-accurate physical material predictions.} 
\label{fig:seg}
\end{figure*}

\section{Experiments}

\subsection{Datasets and Evaluation Protocol}
\noindent \textbf{Datasets.} 
We evaluated the quantitative and qualitative performance using both synthetic and real-captured data from the Amazon Berkeley Objects (ABO) dataset \cite{collins2022abo} and the MVImgNet dataset \cite{yu2023mvimgnet}. Following \cite{zhai2024physical}, we selected 100 validation objects from the ABO dataset. For MVImgNet, we also selected 100 objects. The criterion for selection was to ensure coverage of a diverse range of material categories, and we filtered out cases that could not be accurately classified.
Finally, we manually annotated  detailed material labels for each part of the objects. This process resulted in a final set of 78 labeled cases in the ABO dataset and 100 cases in MVImgNet. Moreover, we also captured 16 objects composed of various materials for robotic grasping. Further details can be found in the Supplementary.

\vspace{0.1cm}
\noindent \textbf{Evaluation protocol.} 
To evaluate the accuracy of material prediction after adding physical properties to 3D Gaussians, we use the mean Intersection over Union (mIoU) metric \cite{everingham2010pascal}. This process involves selecting an angle from which the object can be better observed. The 3D Gaussians render the material information into 2D to form a material segmentation map. Similar to 2D evaluations, we use mIoU as an indicator to assess the accuracy of the material segmentation. For robotic grasping, the Picked-up Rate (PUR) and the No-damage Rate (NDR) evaluate respectively whether objects are picked up without slipping and whether no damages to objects are caused. A final success requires both criteria being met, yielding a final Success Rate (SR).

\subsection{Implementation Details} 
For each object, we collected 30 views with camera centers randomly distributed over a hemisphere around the object. We used 3D Gaussian Splatting for 3D reconstruction, following the default parameter settings. Our model was trained for 5 minutes on a single NVIDIA RTX-A6000 GPU. To accelerate the part-level segmentation and property matching process, we selected only 10 views. For multi-modal model processing, we used GPT-4V as the large multimodal model. For dynamic simulation, we implemented Physgaussian~\cite{xie2023physgaussian} with assigning estimated materials for each 3D Gaussian.
In robot grasping experiments, we utilized a 
\iftoggle{cvprfinal}{Jacobi.ai JSR-1 }{*** (hidden for review) }
robot platform 
\iftoggle{cvprfinal}{\cite{qiu2024ovmm} }{}
equipped with a 
\iftoggle{cvprfinal}{TEK CTAG2F90-C }{*** (hidden for review) }
robotic gripper that has a maximum grasping force up to $40N$. The force-bearing surface at the tip of the gripper is measured to encompass an area of $ A = 0.00011\text{m}^2 $. And a maximum allowable bending curvature $\kappa_{\max} = 0.5$ is used. The robotic gripper's grasping force has been calibrated with its normalized input $15 \leq N_{GF} \leq 100$ before experiment.





\begin{table}[t]
\centering
\caption{Ablation study of Global-to-Local Knowledge Integration and Frequency-Based Voting.}
\label{tab:ablation}
\def\arraystretch{1.1}
\scalebox{0.9}{
\begin{tabular}{cccc}
    \toprule
    \small
    Global-to-local & Voting & Average mIoU (\%$\uparrow$) \\
    \midrule
     & \checkmark &  22.17\\
    \checkmark & & 51.28\\
    \checkmark & \checkmark & \textbf{55.83}\\
    \bottomrule
\end{tabular}
}
\end{table}

\subsection{Material Segmentation.} 
We compared material segmentation performance with the recent work Nerf2Physics \cite{zhai2024physical}, we present both qualitative and quantitative comparisons in Figure \ref{fig:seg} and Table \ref{table:material}. Our method significantly outperforms Nerf2physcis on both synthetic and real-captured data. We also conducted mass and hardness estimation as Nerf2Physics. More results can be found in the Supplementary.

\subsection{Generative Dynamics} \label{dynamics}
\begin{figure}[ht]
\begin{center}
\includegraphics[width=1.0\linewidth]{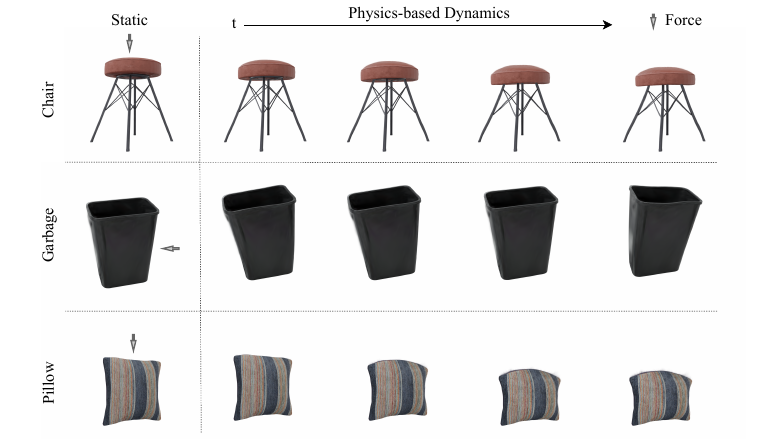}
\end{center}
\vspace{-0.2cm}
   \caption{\textbf{Generative Dynamics}. We present a potential downstream task of 3D Gaussians with physical property, i.e., the generative dynamics. By imposing force, the 3D Gaussians generate corresponding motion. For example, in the first row, we applied a top-down force, the chair exhibited a movement corresponding to the applied force.} 
\label{fig:dynamic}
\end{figure}

Physical simulation is a crucial application of our method because it allows us to directly add all predicted physical properties to the Gaussian points without the need for manual querying and annotation. This integration speeds up dynamic rendering significantly. Figure \ref{fig:dynamic} illustrates some examples showing that the physical properties predicted by our approach can be directly applied in simulation.

\subsection{Robot Grasping} 
To evaluate the effectiveness and performance of our proposed method, we collect 16 objects composed of diverse materials, and implemented three robot grasping baselines with fixed grasping forces, which are widely adopted force-sensitive grasping strategies in robotics. Table \ref{tab:robot_grasping_results} shows our method on material-sensitive grasping with \textit{GaussianProperty} outperforms all the baselines. Several sample cases are shown in Figure \ref{fig:robot_grasping_samples}. Full object list and experiment results can be found in the Supplementary.

\begin{figure}
    \centering
    \includegraphics[width=1.0\linewidth]{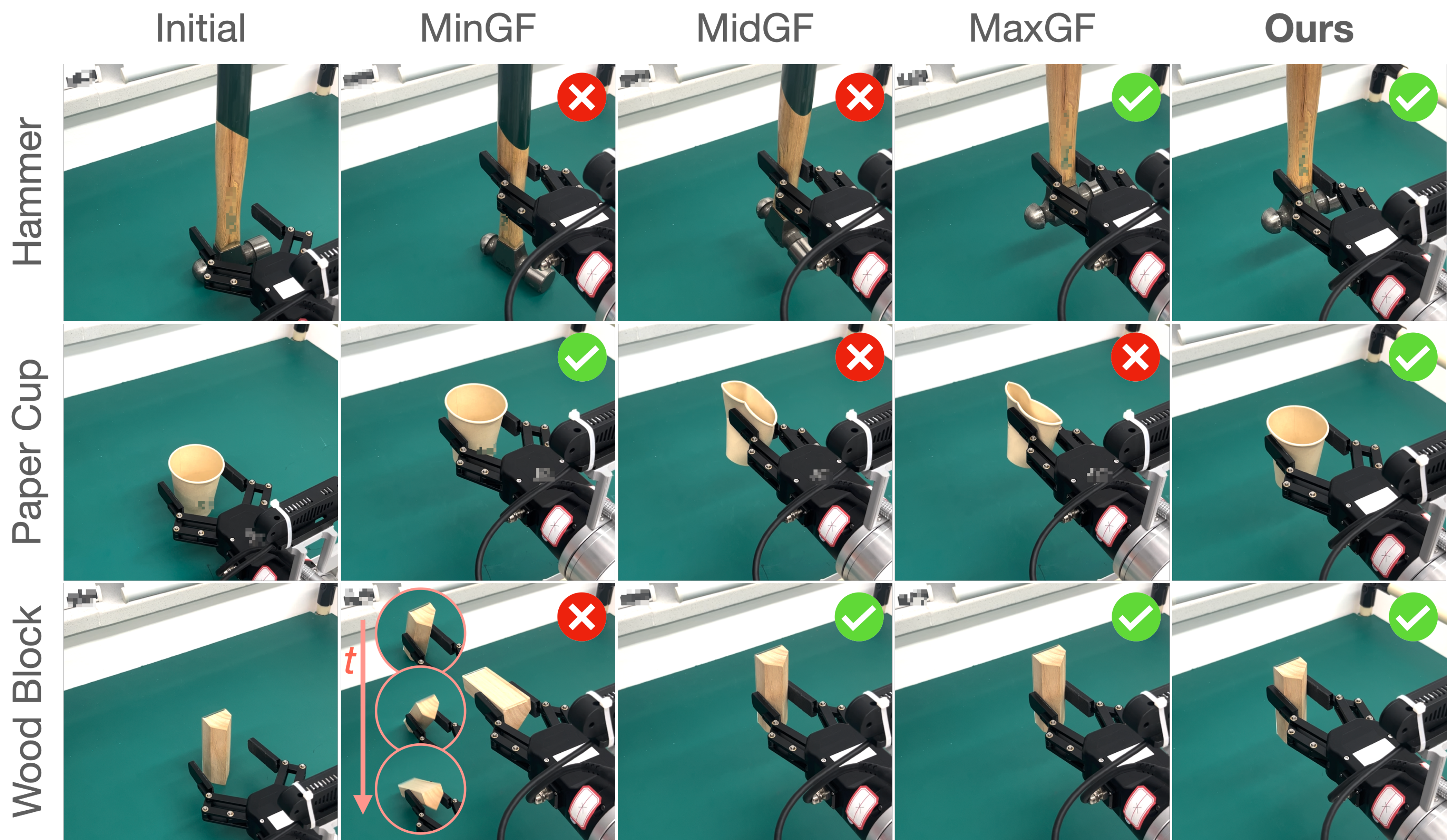}
    \caption{\textbf{Robot Grasping} is a downstream application of \textit{GaussianProperty}. Several sample cases from robot grasping experiments are presented, where we compare our proposed method (right) against three baselines (middle columns), starting from initial configurations (left).}
    \label{fig:robot_grasping_samples}
\end{figure}

\begin{table}[h]
\caption{Results of robot grasping experiments on 16 objects. MinGF, MidGF and MaxGF are baselines with minimum ($N_{GF}=15$), medium ($N_{GF}=60$) and maximum ($N_{GF}=100$) grasping forces applied by the robotic gripper. {\bf Bold}: best results.}
\label{tab:robot_grasping_results}
\small
\centering
\def\arraystretch{1.1}
\scalebox{0.9}{
\begin{tabular}{lcccc}
\toprule
Method & PUR (\%$\uparrow$) & NDR (\%$\uparrow$) & SR (\%$\uparrow$) \\
\midrule
MinGF &  50.00 & \textbf{100.00} &  50.00 \\
MidGF &  87.50 &  81.25 &  68.75 \\
MaxGF & \textbf{100.00} &  75.00 &  75.00 \\
Ours* & \textbf{100.00} & \textbf{100.00} & \textbf{100.00} \\
\bottomrule
\end{tabular}
}
\vspace{-16pt}
\end{table}

\subsection{Ablation Study}

\noindent \textbf{Global-to-Local Knowledge Utilization.}
Table \ref{tab:ablation} demonstrates the impact of incorporating global-to-local knowledge in material segmentation. Without this module, the method only utilizes images of each individual local part of the object for material querying. In contrast, with global-to-local knowledge, the method benefits from a broader context, enabling it to more accurately segment and classify materials. This approach enhances the understanding of the object’s overall structure and finer details, leading to more precise predictions of materials.

\vspace{0.1cm}
\noindent \textbf{Frequency-based Voting Strategy.}
Table \ref{tab:ablation} demonstrates that implementing a frequency-based voting strategy can improve the accuracy of property estimation. By projecting onto multi-view images, we can identify the most frequently occurring material for each part. This frequency-based approach ensures consistency and reliability in the predicted properties by effectively aggregating information from different viewpoints, minimizing errors, and enhancing overall prediction accuracy. We provide an example to demonstrate the effectiveness of the frequency-based voting strategy in the Supplementary.

\section{Conclusion and Limitation}

\noindent \textbf{Limitation}
Despite the promising result of our method on 2D material segmentation, our method struggles to distinguish surface with ambiguous materials. We show an example in the Supplementary.

\noindent \textbf{Conclusion}
In this paper, we explore the issue of estimating physical properties for 3D models, a topic that serves as a foundation for  various downstream task like AR, robotics and simulation, yet remains under-explored. 
The inherent ambiguity and the challenge of acquiring labeled ground-truth data can significantly hinder the estimation. 
Our method, \textit{GaussianProperty}, effectively addresses this challenge by leveraging the recognition capability of large multimodality models and segmentation capability of SAM to achieve a combined global-local reasoning module on 2D space. Then, a voting strategy is employed to project the 2D material property estimation results to 3D Gaussians, a effective and efficient 3D representation, supporting multi-view reconstruction and real-time rendering. We show two potential downstream applications, i.e., physics-based dynamic simulation and robotic grasping. 
Extensive experiments on manually annotated material segmentation dataset and real-world robot grasping experiments validate the effectiveness of the methods we propose.

{
    \small
    \bibliographystyle{arxiv}
    \bibliography{arxiv}
}


\clearpage
\setcounter{page}{1}
\maketitlesupplementary

\appendix

\section{Derivation of Grasping Force}\label{apx:grasping_force_derivation}

In the derivation below, we assume physical properties are uniform distributed over the entire object to grasp, which can be easily extended to more generic situations.

The lower bound of the grasping force $F_{\min}$ is the minimal sufficient force applied on the gripper to lift the object without slipping.
\[
m g \cos\theta = \mu (2 F_{\min} + m g \sin\theta)
\]
\begin{align*}
F_{\min}
&= \frac{1}{2} m g \left( \frac{\cos\theta}{\mu} - \sin\theta \right) \\
&= \frac{1}{2} \rho V g \left( \frac{\cos\theta}{\mu} - \sin\theta \right)
\end{align*}
where $m$, $\rho$ and $V$ are the mass, the density and the volume of the object respectively, $\theta$ is the lifting angle of the gripper with upward at $0$ degree, $\mu$ is the friction coefficient between the gripper finger tips and the object surface, and $g \approx 9.8 m/s^{2}$ is the gravity constant.

The upper bound of the grasping force $F_{\max}$ is the maximal force that does not cause any damage resulted by exceeding the yield stress $\sigma_{y}$ or any undesirable deformation over some maximum allowable bending curvature $\kappa_{\max}$ of the object. Following the formula of bending stress
\[
\frac{\sigma}{y} = \frac{E}{R}
\]
the corresponding maximum stress applied on the force bearing surface at curvature $\kappa_{\max}$ is
\[
\sigma_{c} = \frac{E y(s)}{R_{\min}} = \frac{1}{2} E d \kappa_{\max}
\]
Therefore, the maximal grasping force
\begin{align*}
F_{\max} 
&= A \sigma_{\max} \\
&= \min\left[ A \sigma_{y}, A \sigma_{c} \right] \\
&= \min\left[ A \sigma_{y}, \frac{1}{2} A E d \kappa_{\max} \right]
\end{align*}
where $A$ is the area of the force bearing surface of the object (or equivalently the area of one side of the robot gripper finger tips), $\sigma$ is the bending stress at a point of the object at perpendicular distance $y$ from the neutral axis, $s$ is the outmost point of the force bearing surface, $d$ is the thickness of the force bearing surface of the object, $R = 1 / \kappa$ is the radius of curvature of the neutral axis, and $E$ is Young's modulus of electricity of the object material.

To maximize the grasping reliability, a reasonable choice of grasping force would be $\bar{F} = (F_{\min} + F_{\max}) / 2$. Additionally, the grasping force must be confined within the input bounds of the robotic gripper, and we also attempt to avoid the gripper executing commands close to its input bounds, with preferably $0 \leq \eta \leq 1$ margin. These three principals yield an optimal choice of grasping force 
\[
F^{*} = \begin{cases}
    \left[ \bar{F} \right]_{\left[ F_{\min} \right]_{G} + \eta \Delta F}^{\left[ F_{\max} \right]_{G} - \eta \Delta F} & F_{\min} < F_{\max} \\
    \left[ \bar{F} \right]_{G} & F_{\min} \geq F_{\max}
\end{cases} 
\]
with $\left[ f \right]_{G}$ clipping a force $f$ between the minimum and the maximum grasping forces of robotic gripper $G$, $\left[ f \right]_{f_{\min}}^{f_{\max}}$ clipping $f$ between $f_{\min}$ and $f_{\max}$, and $ \Delta F = \max\left[ 0, \left[ F_{\max} \right]_{G} - \left[ F_{\min} \right]_{G} \right]$. In reality, it is possible to observe $F_{\min} > F_{\max}$, rendering infeasibility to picked up an object without damaging it. And $F^{*}$ remains optimality in such situations.

\section{Robot Grasping Experiment Details}

\begin{figure}
    \centering
    \iftoggle{cvprfinal}{
    \includegraphics[width=0.32\linewidth]{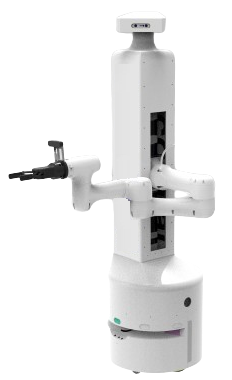}
    }{}
    \includegraphics[width=0.65\linewidth]{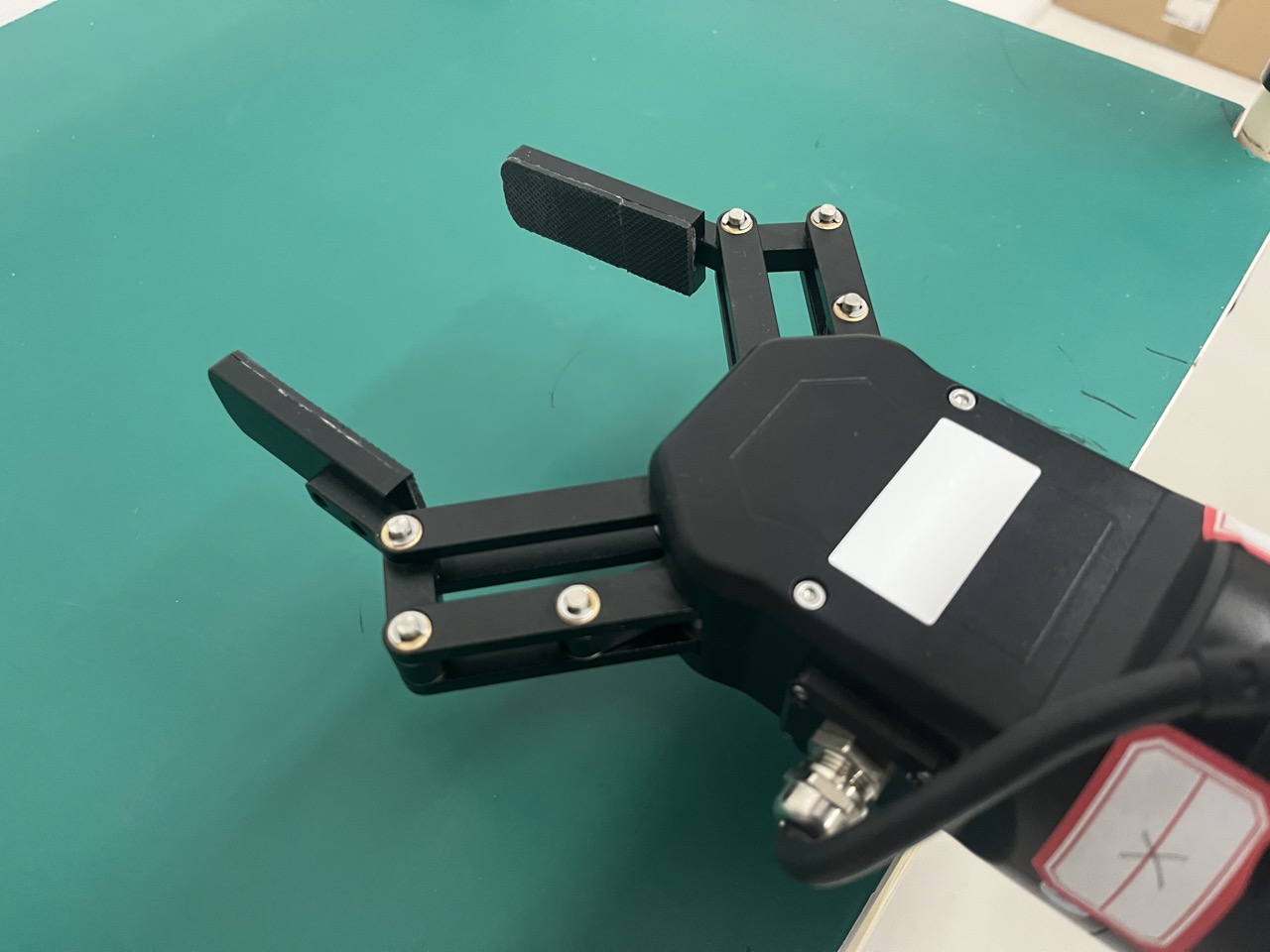}
    \caption{The 
    \iftoggle{cvprfinal}{robot platform (left) and the }{}
    robotic gripper 
    \iftoggle{cvprfinal}{(right) }{}
    utilized in robot grasping experiments.}
    \label{fig:robot_and_gripper}
\end{figure}

In robot grasping experiments, we utilized a 
\iftoggle{cvprfinal}{Jacobi.ai JSR-1 }{*** (hidden for review) }
robot platform 
\iftoggle{cvprfinal}{\cite{qiu2024ovmm} }{}
equipped with a 
\iftoggle{cvprfinal}{TEK CTAG2F90-C }{*** (hidden for review) }
robotic gripper (see Figure \ref{fig:robot_and_gripper}). The force-bearing surface at the tip of the gripper is measured to encompass an area of $ A = 110\text{mm}^2 = 0.00011\text{m}^2 $. And a maximum allowable bending curvature $\kappa_{\max} = 0.5$ is used.

\subsection{Grasping Force Calibration}

\begin{figure}
    \centering
    \includegraphics[width=1.0\linewidth]{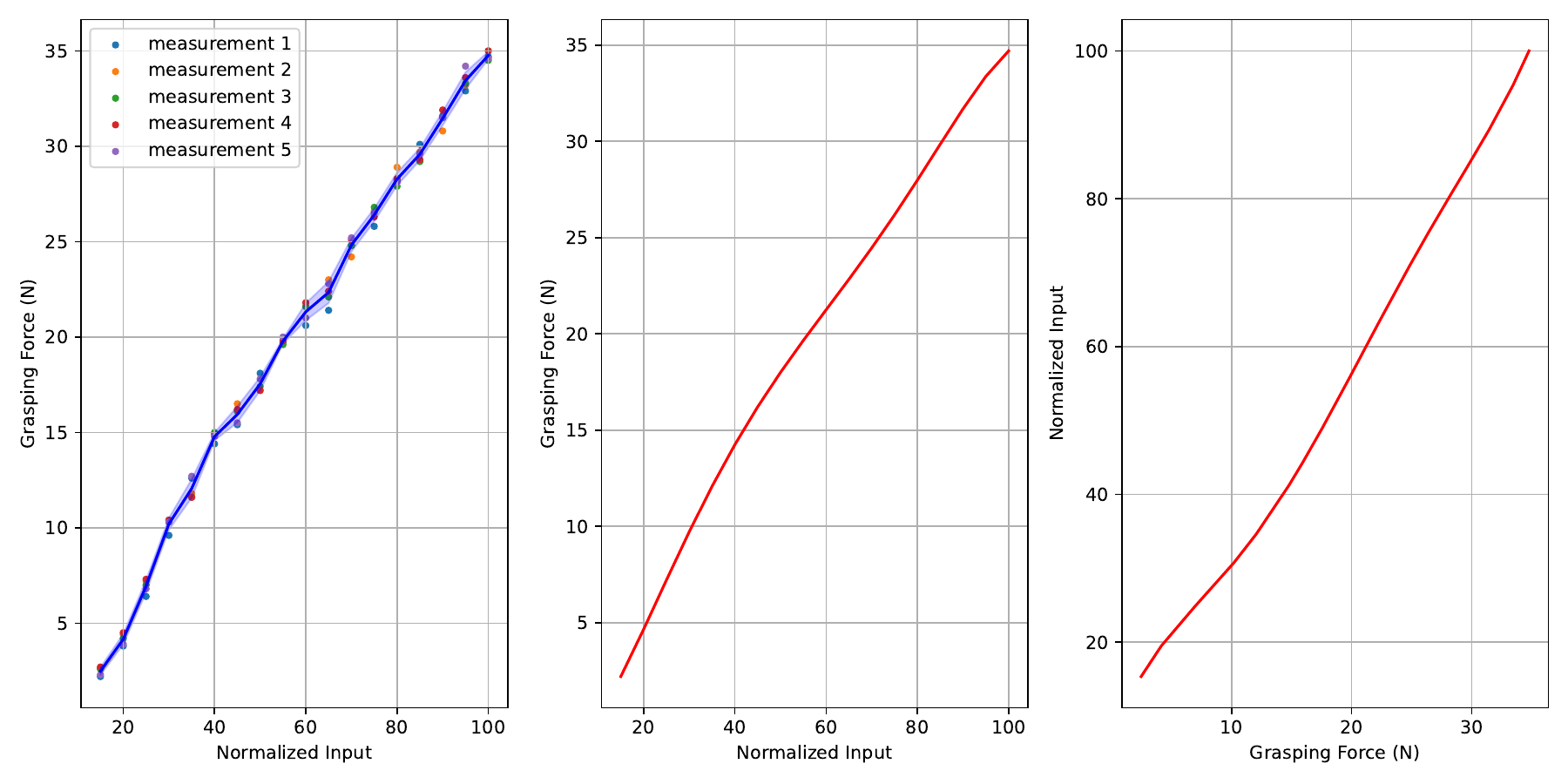}
    \caption{Calibration curve of robotic gripper grasping force (left) and its 5th-order polynomial smoothings (middle and right).}
    \label{fig:gf_cali_curve}
\end{figure}

The robotic gripper employed in this study offers the capability to specify the grasping force on a normalized scale $0 \leq N_{\text{GF}} \leq 100$. Prior to conducting the grasping experiments, we performed a calibration on its grasping force, where 5 measurements are taken for each normalized input data point. The calibration curve is shown in Figure \ref{fig:gf_cali_curve}. We also note that there is a minimum enabling normalized input, and the robotic gripper is only enabled with normalized input $N_{\text{GF}} \geq 15$.

\subsection{Full Object List and Experiment Results}

\begin{figure}[ht]
    \centering
    
    \includegraphics[width=1.0\linewidth]{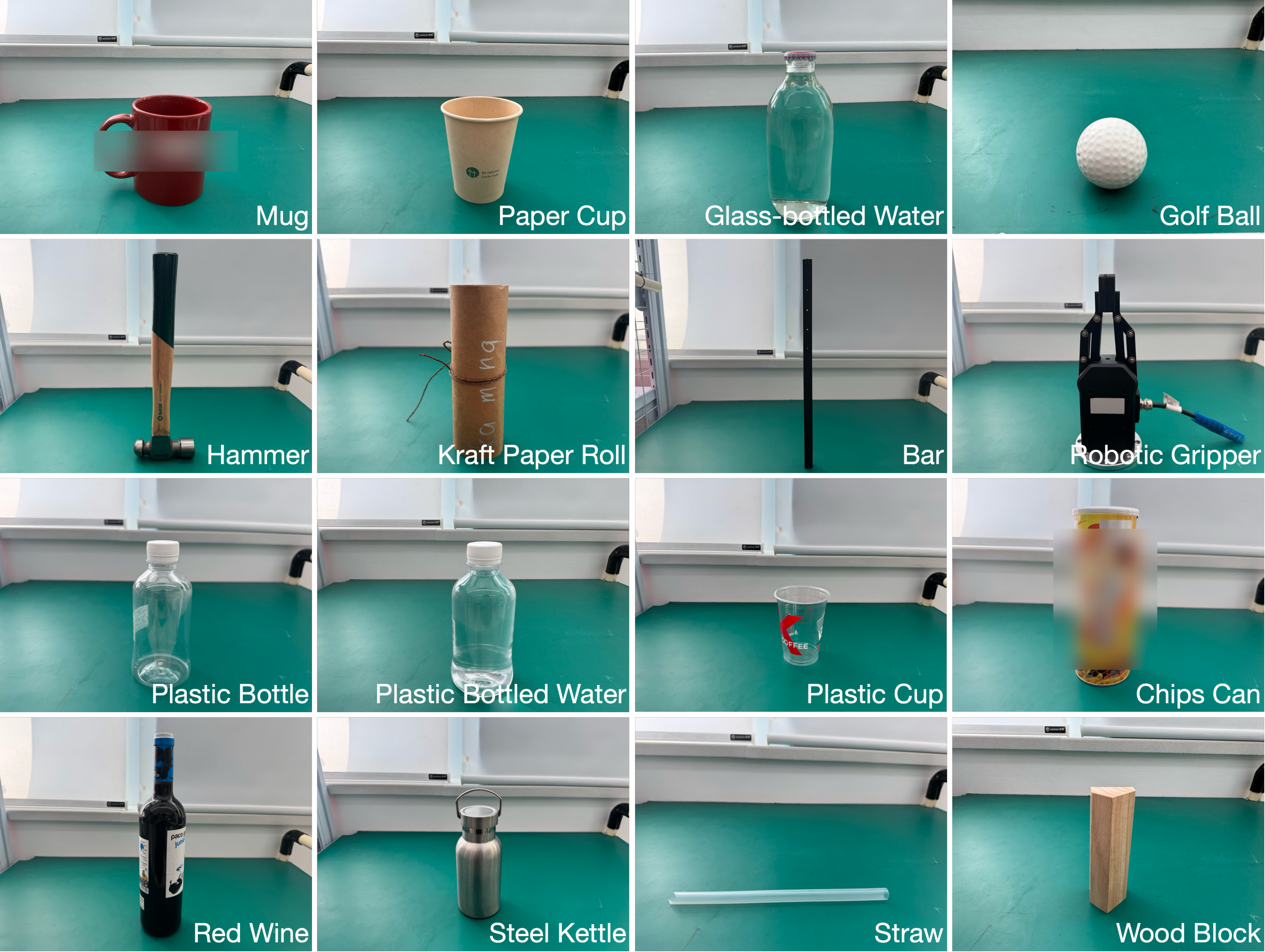}
    
    \caption{List of selected objects for robot grasping experiments.}
    \label{fig:mpg_objects}
\end{figure}

We collected real-world 16 objects for the robot grasping experiments, as illustrated in Figure \ref{fig:mpg_objects}. This collection represents a diverse range of weights and materials, including plastic, ceramic, paper, steel, wood, and glass, etc.
These objects are commonly encountered in everyday life, and the material properties of their different parts exhibit significant variability. Consequently, naive grasping strategies that do not account for material adaptability may find struggling to grasp all of these items in an effective and safe manner.

\begin{figure*}[ht]
\centering
\includegraphics[width=1\textwidth]{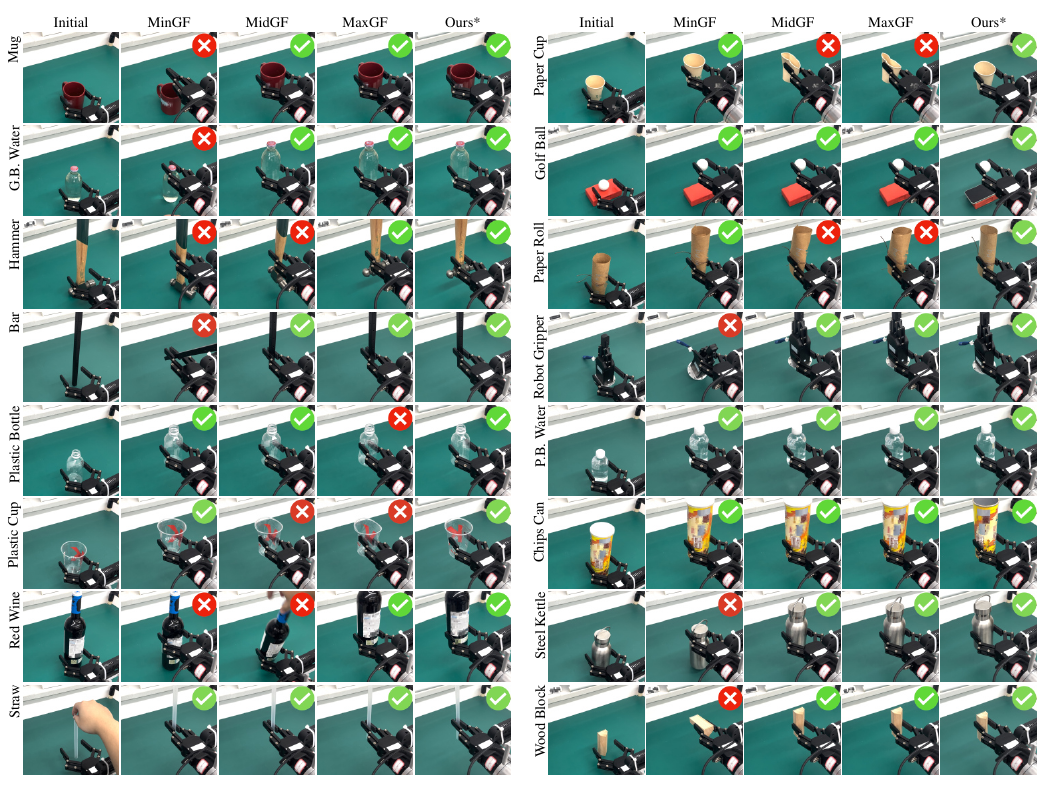}
   \caption{\textbf{Complete robot grasping experiment results.} The 16 test cases along with results in robot grasping experiments are listed. We compare our proposed method (right) against three baselines (middle columns), starting from initial configurations (left). \textbf{You can view the MP4 videos of the experiments in our project page.}} 
\label{fig:grasping_comparison_full}
\end{figure*}

We compare our proposed method on integrating \textit{GaussianProperty} to material-sensitive robot grasping with three baselines, namely MinGF (with the minimum grasping force, $N_{GF}=15$), MidGF (with medium grasping force, $N_{GF}=60$) and MaxGF (with maximum grasping force, $N_{GF}=100$).
Table \ref{tab:robot_grasping_results} in the main PDF listed the detailed experiment results. As summarized in Table \ref{tab:robot_grasping_results}, our method outperforms all the baselines and achieves a success rate of 100\% on all the test objects, by successfully picking them up without slippery or causing any damage or undesirable deformation to them. Figure \ref{fig:grasping_comparison_full} shows the results of the complete robot grasping experiment.

\section{More Results of Experiments}

\subsection{Datasets}
For mass estimation, we use the ABO dataset, which provides mass data for each object. Since the NeRF2Physics method does not include a corresponding hardness dataset, we constructed our own dataset for hardness estimation using a similar methodology. Our dataset includes 10 household items, each captured in a realistic home setting. It features multi-view images paired with Shore hardness measurements. We captured the images and their corresponding poses with an iPhone 13 camera. For each object, hardness was measured at 10 specific points using a hardness tester, with each measurement averaged over three trials. Each measurement point is annotated with pixel coordinates in the images. Notably, Shore A and Shore D hardness testers use different indenters: Shore A measures within a range of 0-100, while Shore D spans a range of 100-200.

\subsection{Evaluation Metrics}
We report the following metrics, where $p$ is the ground-truth mass/hardness and $\hat{p}$ is the estimated mass/hardness:
\begin{itemize}[leftmargin=*]
    \item Absolute difference error (ADE): $|p - \hat{p}|$,
    \item Absolute log difference error (ALDE): $|\ln p - \ln \hat{p}|$,
    \item Absolute percentage error (APE): $\left|\frac{p - \hat{p}}{p}\right|$,
    \item Min ratio error (MnRE): $\min\left(\frac{p}{\hat{p}}, \frac{\hat{p}}{p}\right)$, and
    \item Pairwise Relationship Accuracy (PRA): 
    \[
    \text{PRA} = \frac{1}{N_\text{pairs}} \sum_{i \neq j} \mathbb{1}\big((p_i > p_j) \iff (\hat{p}_i > \hat{p}_j)\big),
    \]
    where $N_\text{pairs}$ is the total number of object pairs, and $\mathbb{1}(\cdot)$ is the indicator function, which equals 1 if the condition inside is true, and 0 otherwise.
\end{itemize}

\subsection{Hardness Estimation}
\begin{figure}[ht]
\centering
\includegraphics[width=1\linewidth]{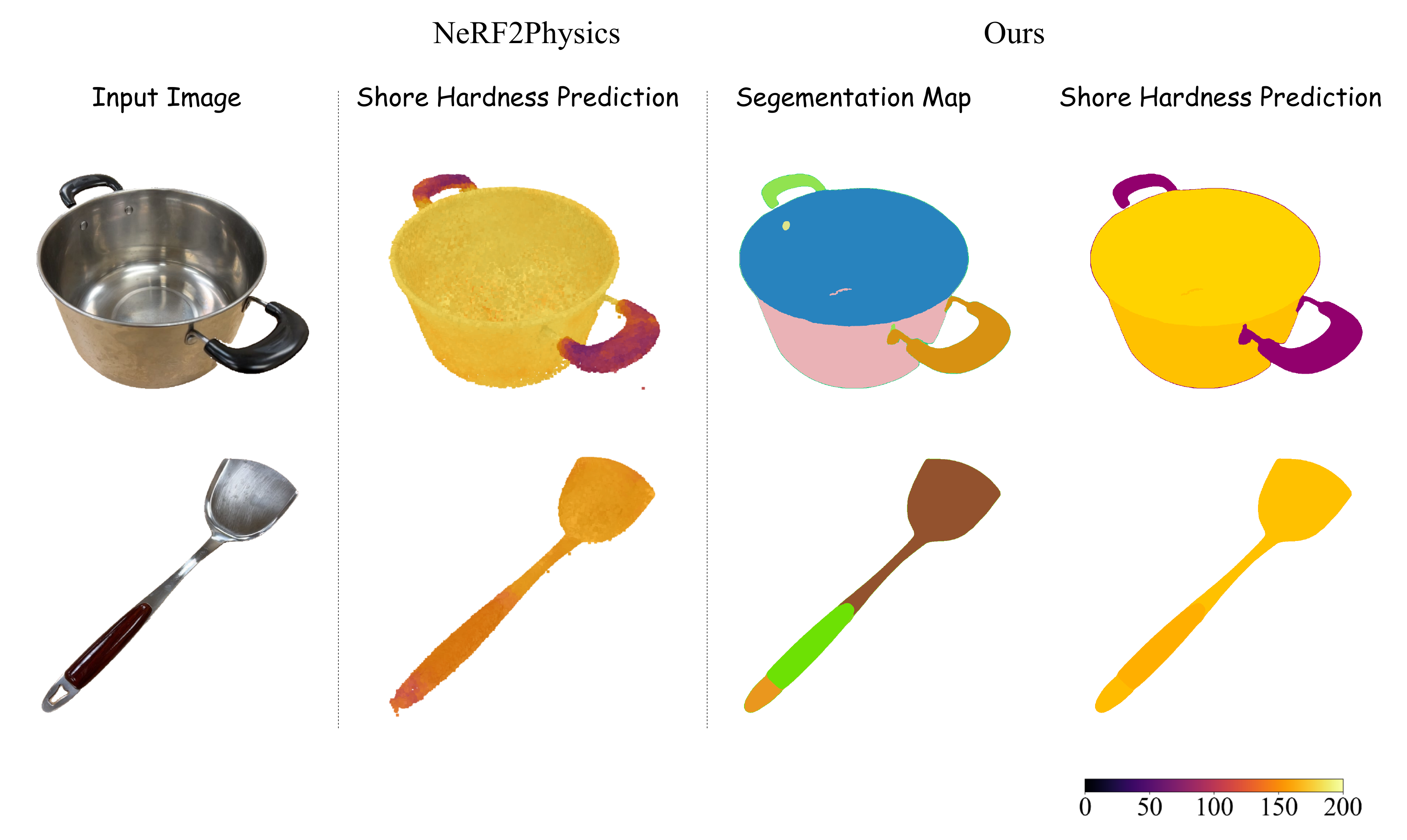}
   \caption{\textbf{Qualitative comparison of hardness prediction}. Compared to NeRF2Physics, our method provides more accurate hardness prediction with clear boundaries.} 
\label{fig:hardnss}
\end{figure}

Table \ref{tab:hardness_results} presents the quantitative results of our method and NeRF2Physics on the hardness estimation task. Our approach outperforms NeRF2Physics across all metrics, demonstrating a significantly improved capability to accurately assess object attributes. This improvement can be attributed to the integration of LMMs, our method can have a more accurate understanding of each part of the object and form an accurate and clear-cut hardness estimation. Figure \ref{fig:hardnss} illustrates the hardness estimation results produced by our method on the same case without the application of voting .

\begin{table}[ht]
\caption{Estimation of per-point Shore hardness on the real-captured in-house collected dataset (10 objects, 100 points). {\bf Bold}: best model. }
\small
\centering
\setlength\tabcolsep{1.0pt}
\def\arraystretch{1.1}
\scalebox{0.88}{
\begin{tabular}{lccccc}
\toprule
Method  & ADE ($\downarrow$) & ALDE ($\downarrow$) & APE ($\downarrow$) & MnRE ($\uparrow$) & PRA ($\uparrow$)  \\
\midrule

NeRF2Physics & 35.917 & 0.328 &  0.294 & 0.748 & 0.575\\

Ours* & \textbf{28.583} & \textbf{0.220} & \textbf{0.198} & \textbf{0.820} & \textbf{0.686}\\

\bottomrule
\label{tab:hardness_results}
\end{tabular}
}
\vspace{-16pt}
\end{table}

\subsection{Mass Estimation}
3D Gaussian object reconstruction allows for the estimation of the volumes of various parts composing an object. By integrating this with material property prediction where densities of different object parts are inferred, we can derive an overall estimation of object mass. We subsequently compare our mass estimation with the baseline NeRF2Physics, Table \ref{tab:mass} demonstrating that our method provides more accurate quality assessments and significantly outperforms the baseline across most indicators.

\begin{table}[ht]
\caption{Mass estimation on ABO dataset. {\bf Bold}: best results.}
\small
\centering
\def\arraystretch{1.1}
\setlength\tabcolsep{1.5pt}
\scalebox{0.9}{
\begin{tabular}{lcccc}
\toprule
Method  & ADE ($\downarrow$) & ALDE ($\downarrow$) & APE ($\downarrow$) & MnRE ($\uparrow$)  \\
\midrule
NeRF2Physics & 12.761 & 0.803 & \textbf{0.589} & 0.498 \\

Ours* & \textbf{5.960} & \textbf{0.744} & 1.609 & \textbf{0.559} \\

\bottomrule
\label{tab:mass}
\end{tabular}
}
\vspace{-16pt}
\end{table}

\section{Additional details of Our Method}

\subsection{Segmentation Process Using SAM at Different Levels}

\begin{figure}[ht]
\begin{center}
\includegraphics[width=1.0\linewidth]{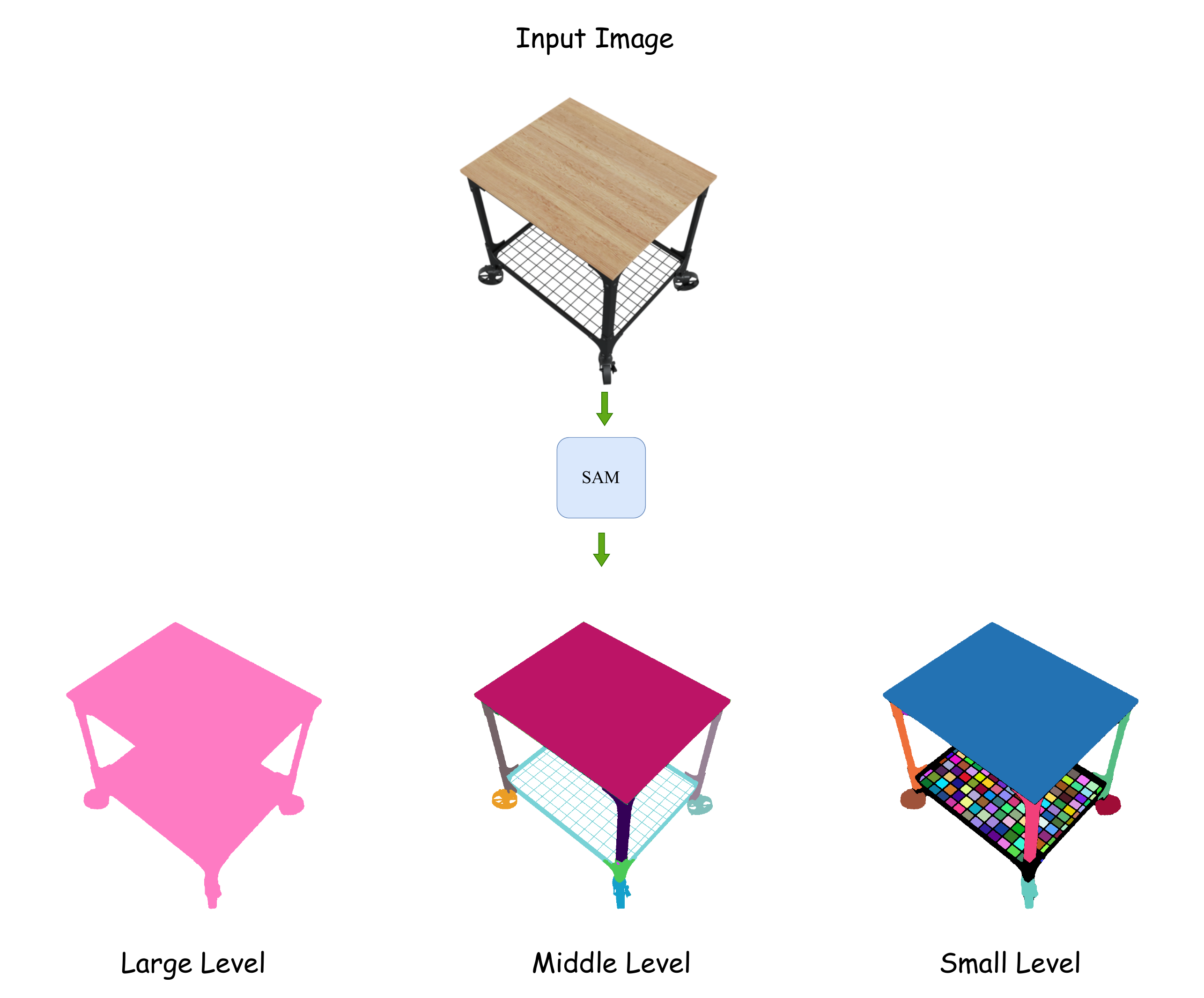}

\end{center}
   \caption{\textbf{Segmentation process using SAM at different levels} of granularity. From left to right: the input image, large-level segmentation, middle-level segmentation, and small-level segmentation. For our model, we selected the middle-level of SAM prediction to balance part-level object understanding and computational efficiency.} 
\label{fig:sam_level}
\end{figure}

We employ the Segment Anything Model (SAM) to generate segmentations at three levels of granularity: large-level, middle-level, and small-level (Figure~\ref{fig:sam_level}). Large-level segmentation simplifies object grouping but lacks detail, while small-level segmentation captures fine details at the cost of increased computational complexity. To balance object understanding and efficiency, we select the middle-level segmentation, which preserves meaningful part-level details without excessive fragmentation, making it ideal for our tasks.

\section{Detail of Data Labeling}
\begin{figure}[ht]
\begin{center}
\includegraphics[width=1.0\linewidth]{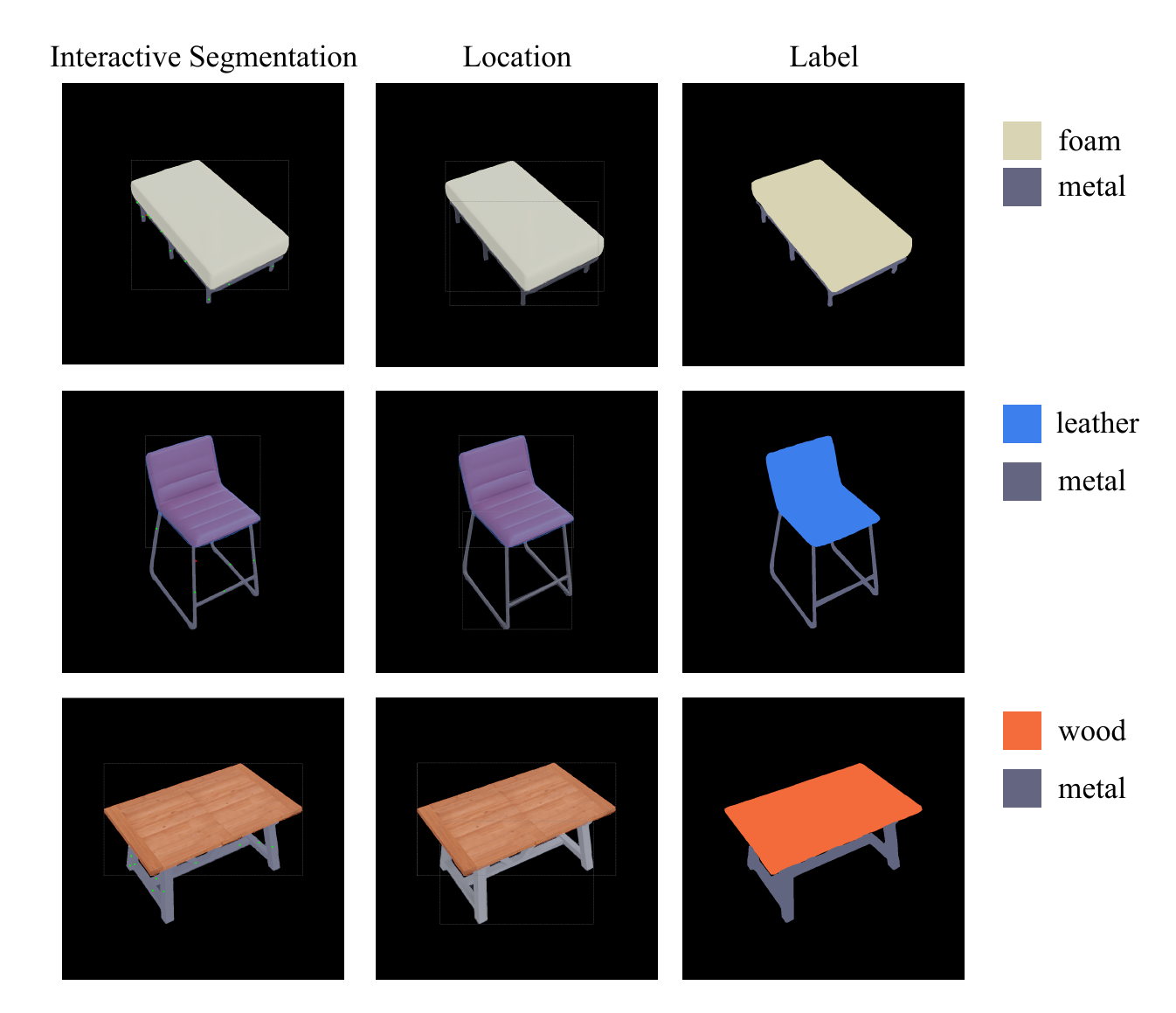}
\end{center}
   \caption{\textbf{Examples of data labeling}. These objects are sourced from the ABO-500 dataset. } 
\label{fig:label_supply}
\end{figure}

We utilize the open-source interactive segmentation tool EISeg \cite{hao2022eiseg} to annotate certain views of each object from ABO and MVImgNet, as shown in Figure \ref{fig:label_supply}. Since some materials are difficult to distinguish by the naked eye, such as aluminum and iron within the metal category. We established ten precise and unambiguous labels for a fair comparison. The labels are: wood, metal, plastic, glass, fabric, foam, marble, ceramic, concrete, and leather.

\subsection{Prompting Details}

\begin{figure}[ht]
\begin{center}
\includegraphics[width=1.0\linewidth]{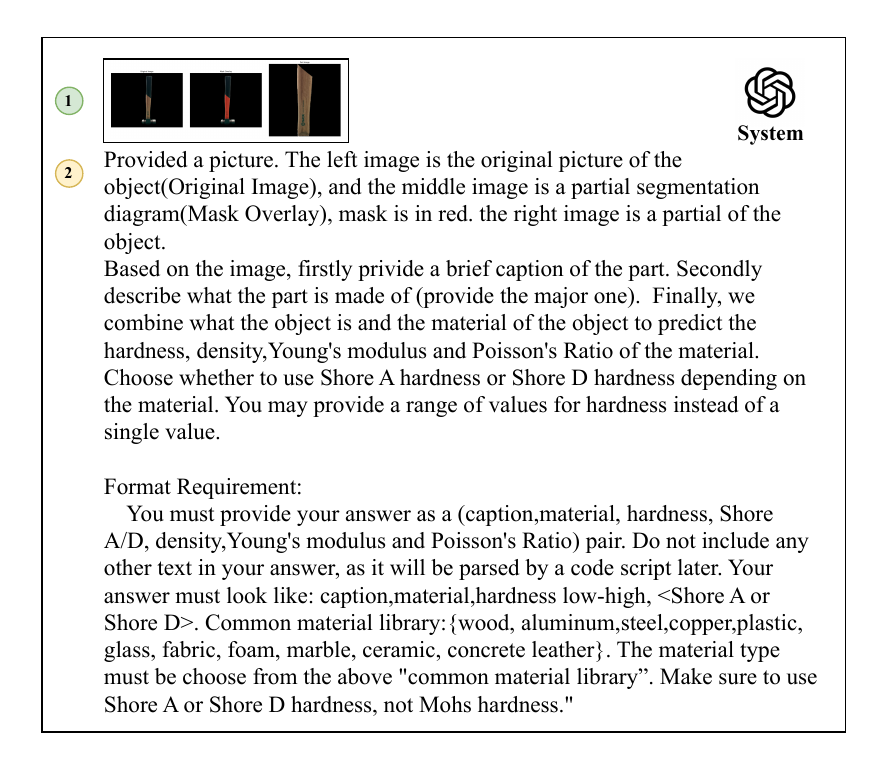}

\end{center}
   \caption{Prompt used for proposing materials and other physical properties.} 
\label{fig:system_prompt}
\end{figure}

We provide the prompts used for material proposal with other physical propertie such as hardness, density,Young's modulus and Poisson's Ratio in Figure \ref{fig:system_prompt}.

\subsection{Effects of Frequency-based Voting Strategy}
\begin{figure}[ht]
\begin{center}
\includegraphics[width=1.0\linewidth]{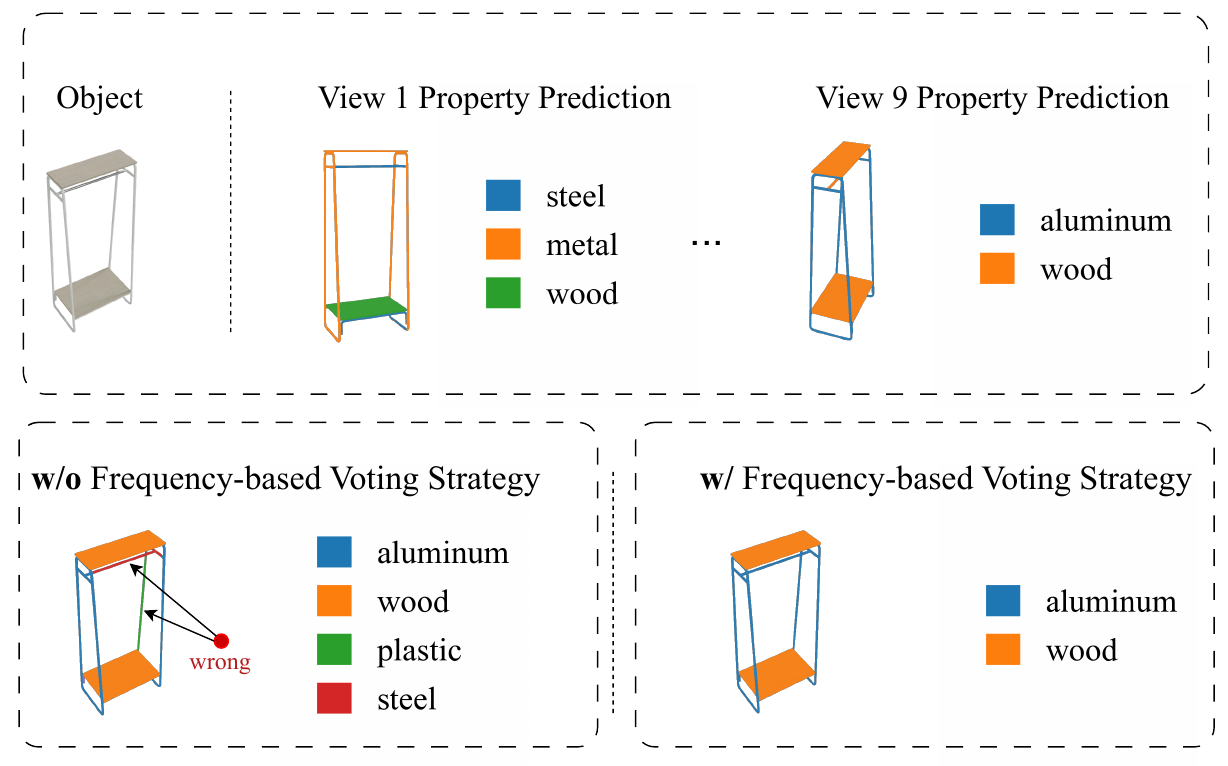}
\end{center}
   \caption{\textbf{Effects of Frequency-based Voting Strategy}. We provide an example to demonstrate the effectiveness of the frequency-based voting strategy. The result misclassified the ``aluminum'' and ``wood'' into ``plastic'' and ``'steel' without voting strategy.} 
\label{fig:ablation}
\end{figure}

Figure \ref{fig:ablation} showcases that implementing a frequency-based voting strategy can enhance the accuracy of property estimation. By projecting to multi-view images, we can determine the most frequently occurring material for each part. This frequency-based approach ensures consistency and reliability in the predicted properties, effectively aggregating information from different viewpoints, minimizing errors and improving overall prediction accuracy.

\section{More qualitative results of Material Segmentation}

\begin{figure*}
\begin{center}
\includegraphics[width=0.9\textwidth]{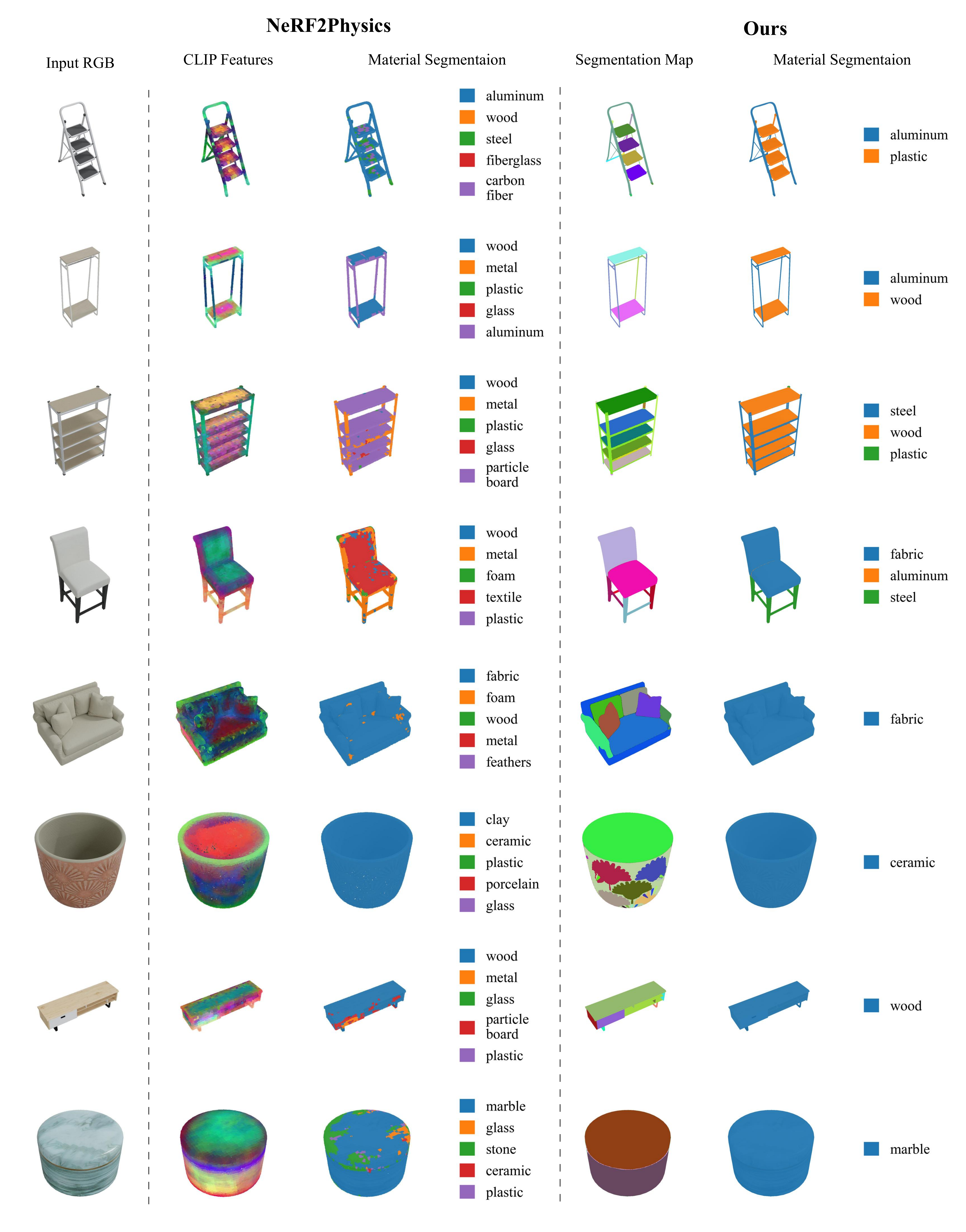}
\end{center}
   \caption{\textbf{Qualitative comparison of Material Segmentation}. These objects are sourced from the ABO-500 dataset. } 
\label{fig:seg_supply}
\end{figure*}

\begin{figure*}
\begin{center}
\includegraphics[width=0.65\textwidth]{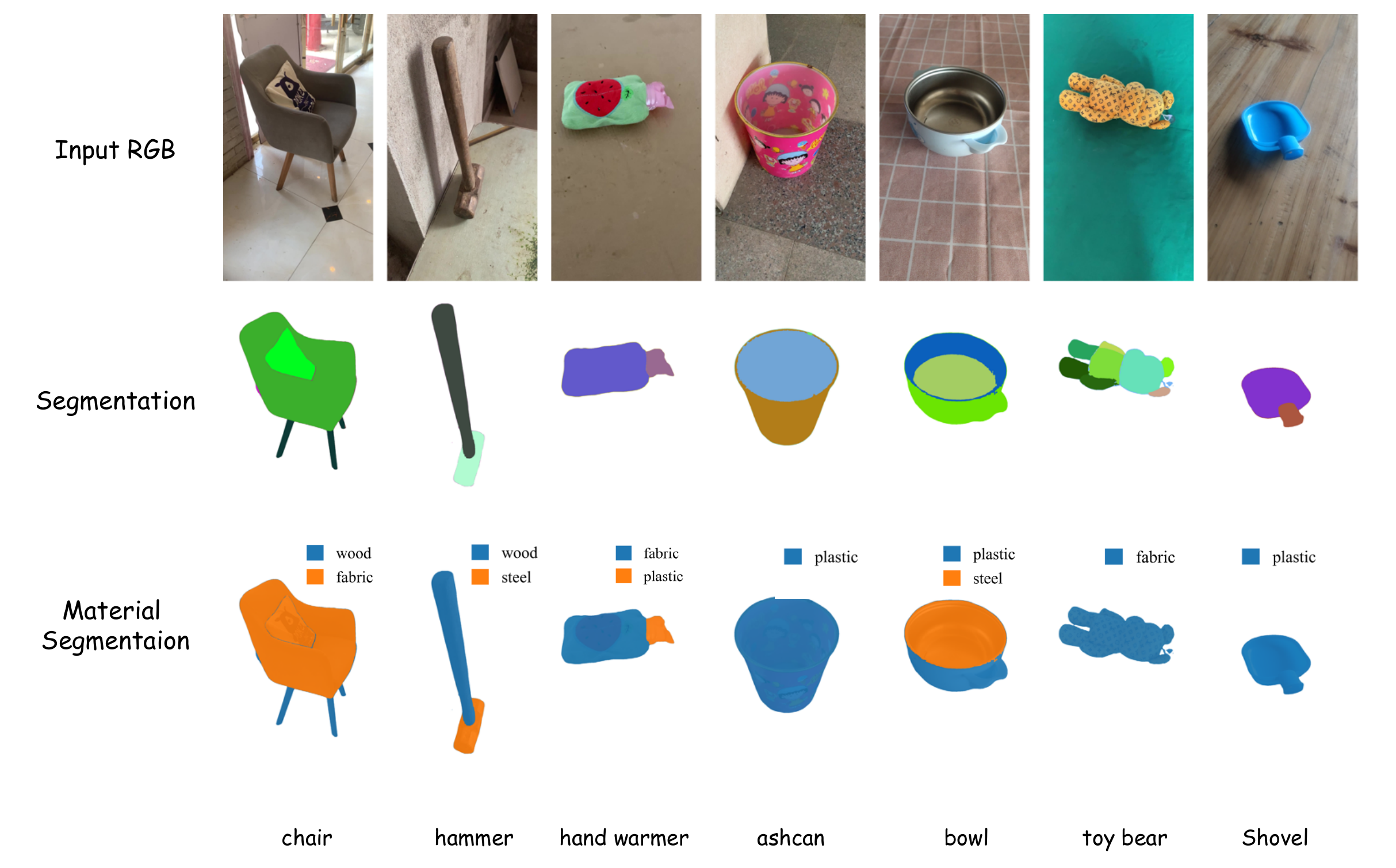}
\end{center}
   \caption{\textbf{Qualitative results of object material segmentation} on MVImgNet. Our model makes reasonable and boundary-accurate  material predictions for objects with multiple or single materials.} 
\label{fig:MVImgNet}
\end{figure*}

In the supplementary material, we provide additional performance comparisons with the baseline model Nerf2Physics. As shown in Figure \ref{fig:seg_supply}, our method predicts the physical properties of objects more accurately. We also show some cases on MVImgNet dataset in Figure \ref{fig:MVImgNet}.

\section{Failure cases}

\begin{figure*}
\begin{center}
\includegraphics[width=0.65\textwidth]{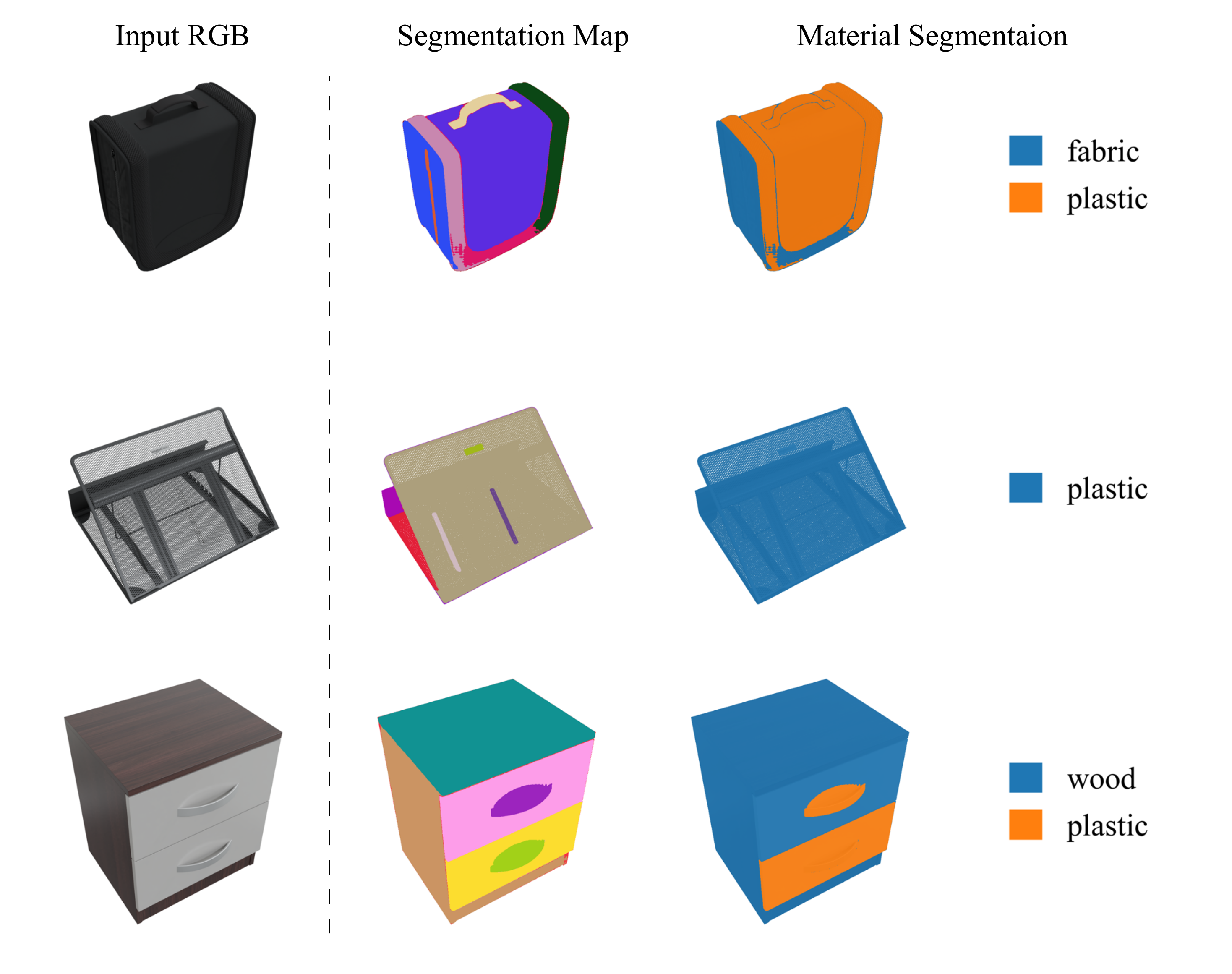}
\end{center}
   \caption{\textbf{Examples of Challenging Material Segmentation Cases}. These objects are sourced from the ABO-500 dataset. } 
\label{fig:fail_supply}
\end{figure*}

However, our method still has limitations. For instance, when the surface texture of an object is ambiguous, it can lead to incorrect classification of material categories, as illustrated in Figure \ref{fig:fail_supply}.

\end{document}